\documentclass[DIV12]{scrartcl}

\usepackage{hyperref}
\usepackage[utf8]{inputenc}
\usepackage{tabularx}
\usepackage{setspace}
\usepackage{graphicx}
\usepackage{booktabs}
\usepackage{amsmath}
\usepackage{color}
\usepackage{url}
\def\eg{e.g.,~\@}
\def\ie{i.e.,~\@}

\newcommand\figref{Figure~\ref}

\newcommand\tabref{Table~\ref}

\hypersetup{
	colorlinks=true,
	linkcolor=black,
	citecolor=black,
	urlcolor=black
}

\date{}
\usepackage{authblk}
\usepackage{blindtext}

\usepackage{natbib}
\title{Vision-based Detection of Acoustic Timed Events: a Case Study on Clarinet Note Onsets}

\usepackage{fancyhdr}
\begin{document}

\author[1]{A.\ Bazzica\thanks{alessio.bazzica@gmail.com (now at Google)}}
\author[2]{J.C.~van Gemert}
\author[1]{C.C.S.~Liem}
\author[1]{A.~Hanjalic}

\affil[1]{\small Multimedia Computing Group - Delft University of Technology, The Netherlands}
\affil[2]{\small Vision Lab - Delft University of Technology, The Netherlands}

\maketitle

\thispagestyle{fancy} 

\begin{abstract}
Acoustic events often have a visual counterpart. Knowledge of visual information can aid the understanding of complex auditory scenes, even when only a stereo mixdown is available in the audio domain, \eg identifying which musicians are playing in large musical ensembles. In this paper, we consider  a vision-based approach to note onset detection.
As a case study we focus on challenging, real-world  clarinetist videos  and carry out preliminary experiments on a 3D convolutional neural network based on multiple streams and purposely avoiding temporal pooling.
We release an audiovisual dataset with 4.5 hours of clarinetist videos together with cleaned annotations which include about 36,000 onsets and the coordinates for a number of salient points and regions of interest.
By performing several training trials on our dataset, we learned that the problem is challenging. We found that the CNN model is highly sensitive to the optimization algorithm and hyper-parameters, and that treating the problem as binary classification may prevent the joint optimization of precision and recall.
To encourage further research, we publicly share our dataset, annotations and all models and detail which issues we came across during our preliminary experiments.

\bigskip


\end{abstract}

\section{Introduction}
Acoustic timed events take place when persons or objects make sound, \eg when someone speaks or a musician plays a note.
Frequently, such events also are visible: a speaker's lips move, and a guitar cord is plucked.
Using visual information we can link sounds to items or people and can distinguish between sources when multiple acoustic events have different origins. We then can also interpret our environment in smarter ways: \eg identifying the current speaker, and indicating which instruments are playing in an ensemble performance.

Understanding scenes through sound and vision has both a \textit{multimodal} and a \textit{cross-modal} nature.
The former allows us to recognize events using auditory and visual stimuli jointly. But when \eg observing a door bell button being pushed, we can cross-modally infer that a bell should ring.
In this paper, we focus on the cross-modal case to detect acoustic timed events from video. Through visual segmentation, we can spatially isolate and analyze sound-making sources at the individual player level, which is much harder in the audio domain~\citep{BazzicaCVIU2015}.

As a case study, we tackle the musical note onset detection problem by analyzing clarinetist videos.
Our interest in this problem is motivated by the difficulty of detecting onsets in audio recordings of large (symphonic) ensembles. Even for multi-track recordings, microphones will also capture sound from nearby instruments, making it hard to correctly link onsets to the correct instrumental part using audio alone. Knowing where note onsets are and to which part they belong is useful for solving several real-world applications, like audio-to-score alignment, informed source separation, and automatic music transcription.

Recent work on cross-modal lip reading recognition~\citep{DBLP:journals/corr/ChungSVZ16} shows the benefit of exploiting video for a task that has traditionally been solved only using audio.
In~\citep{Bochen2017}, note onset matches between a synchronized score and a video are used to automatically link audio tracks and musicians appearing in a video.
The authors show a strong correlation between visual and audio onsets for bow strokes. However, while this type of visual onset is suitable for strings, it does not correlate well to wind instruments.
In our work we make an important step towards visual onset detection in realistic multi-instrument settings focusing on visual information from clarinets, which has sound producing interactions (blowing, triggering valves, opening/closing holes) representative for wind instruments in general.

Our contributions are as follows: (i) defining the visual onset detection problem, (ii) building a novel 3D convolutional neural network (CNN) \citep{DBLP:conf/iccv/TranBFTP15} without temporal pooling and with dedicated streams for several regions of interest (ROIs), (iii) introducing a novel audiovisual dataset of 4.5 hours with about 36k annotated events, and (iv) assessing the current gap between vision-based and audio-based onset detection performance.

\section{Related work}\label{sec:relwork}
When a single instrument is recorded in isolation, audio onset detectors can be used.
A popular choice is \citep{DBLP:conf/icassp/SchluterB14}, which is based on learning time-frequency filters through a CNN applied to the spectrogram of a single-instrument recording.
While state-of-the-art performance is near-perfect, audio-only onset detectors are not trained to handle multiple-instrument cases. To the best of our knowledge, such cases also have not been tackled so far.

A multimodal approach~\citep{DBLP:journals/tmm/BarzelayS10} spots independent audio sources, isolates their sounds and is validated on four audiovisual sequences with two independent sources.
As the authors state~\citep{DBLP:journals/tmm/BarzelayS10}, their multimodal strategy is not applicable in crowded scenes with frequent audio onsets. Therefore, it is not suitable when multiple instruments mix down into a single audio track.

A cross-modal approach~\citep{DBLP:conf/nime/BurnsW06} uses vision  to retrieve guitarist fingering gestures.
An audiovisual dataset for drum track transcription is presented in \citep{DBLP:conf/ismir/GilletR06} and
\citep{Dinesh2017} addresses audiovisual multi-pitch analysis for string ensembles. All works devise specific visual analysis methods for each type of instrument, but do not consider transcription or onset detection for clarinets.

Action recognition aims to understand events. Solutions based on 3D convolutions \citep{DBLP:conf/iccv/TranBFTP15} use frame sequences to learn spatio-temporal filters, whereas two-streams networks \citep{DBLP:conf/cvpr/FeichtenhoferPZ16} add a temporal optical flow stream.
A recurrent network \citep{DBLP:conf/cvpr/DonahueHGRVDS15} uses LSTM units on top of 2D convolutional networks.
While action recognition is similar to visual-based acoustic timed events detection, there is a fundamental difference: action recognition aims to detect the presence or absence of an action in a video. Instead, we are interested in the exact temporal location of the onset.

In action localization~\citep{mettes2016spotOn} the task is to find what, when, and where an action happens. This is modeled with a ``spatio-temporal tube'': a list of bounding-boxes over frames. Instead, we are not interested in the spatial location; we aim for the temporal location only, which due to the high-speed nature of onsets reverts to the extreme case of a single temporal point.

\section{Proposed baseline method}\label{sec:method}
\begin{figure*}[h]
	\centering
	\includegraphics[width=1.0\textwidth]{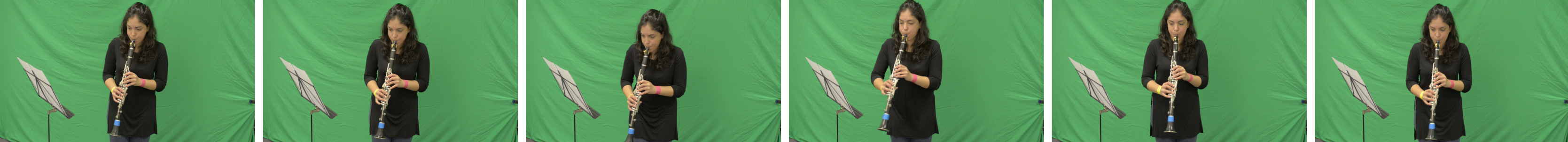}
	\caption{Raw video frames example.}
	\label{fig:raw-input}
\end{figure*}

Together with our dataset, we offer a baseline model for onset detection. The input for our model is a set of sequences generated by tracking a number of oriented ROIs from a video of a single clarinetist (see \figref{fig:raw-input}). For now, as a baseline, we assume that in case of a multi-player ensemble, segmentation of individual players already took place.
The ROIs consider those areas in which the sound producing interactions take place: mouth, left/right hands, and clarinet tip, since they are related to blowing, fingering, and lever movements respectively.

\begin{figure*}[h]
	\centering
	\includegraphics[width=1.0\textwidth]{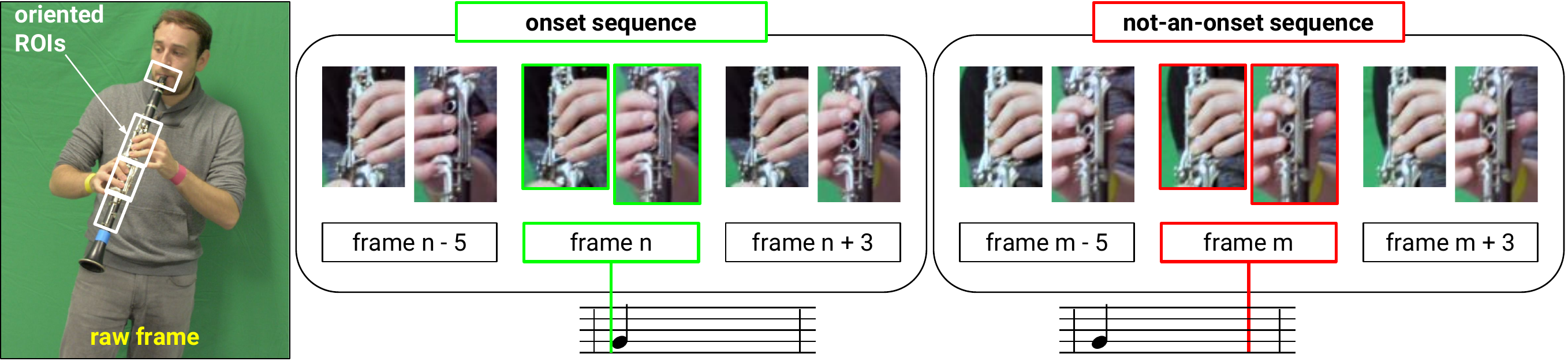}
	\caption{Onset and not-an-onset input sequence examples with 2 ROIs from 3 frames.}
	\label{fig:examples}
\end{figure*}

Each sequence is labeled by determining if a note has started during the time span of the \textit{reference frame}.
A sequence consists of 5 preceding frames, the reference frame, and 3 succeeding frames, forming a sequence of 9 consecutive frames per ROI.
We use a shorter future temporal context because the detector may otherwise get confused by \textit{anticipation} (getting ready for the next note).
Examples of onset and not-an-onset inputs are shown in \figref{fig:examples}.

\begin{figure*}
	\centering
	\includegraphics[width=1.0\textwidth]{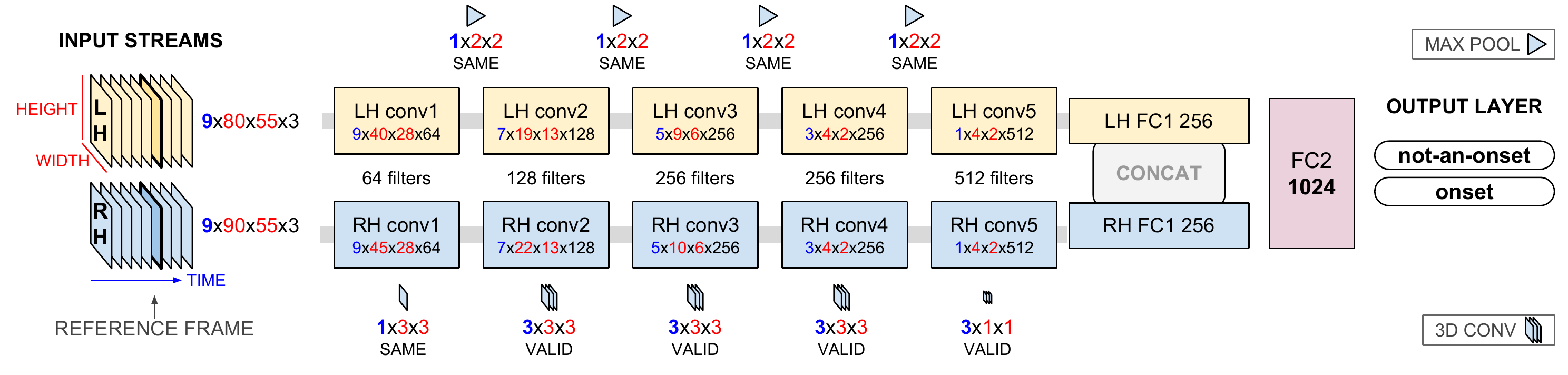}
	\caption{Proposed model based on 3D CNNs, slow fusion, and multiple streams (one for each ROI). LH and RH indicate the left and right hand streams respectively.}
	\label{fig:architecture}
\end{figure*}

Our model relies on multiple \textit{streams}, one for each ROI.
Each stream consists of 5 convolutional layers (CONV1-5), with a fully-connected layer on top (FC1).
All the FC1 layers are concatenated and linked to a global fully-connected layer (FC2).
All the layers use ReLU units.
The output consists of two units (``not-an-onset'' and ``onset'').
\figref{fig:architecture} illustrates our model and, for simplicity, it only shows one stream for the left hand and one for the right one.

To achieve the highest possible temporal resolution, we do not use temporal pooling.
We use spatial pooling and padding parameters to achieve slow fusion throughout the 5 convolutional layers.
We aim to improve convergence and achieve regularization using batch normalization (BN) \citep{DBLP:conf/icml/IoffeS15}, L2 regularization and dropout.
Since we use BN, we omit the bias terms in every layer including the output layer.

We use weighted cross-entropy as loss function to deal with the unbalanced labels (on average, one onset every 15 samples).
The loss is minimized using the RMSprop algorithm.
While training, we shuffle and balance the mini-batches.
Each mini-batch has 24 samples, half of which are not-an-onset ones, 25\% onsets and 25\% \textit{near-onsets}, where a near-onset is a sample adjacent to an onset.
Near-onset targets are set to $(0.75, 0.25)$, \ie the non-onset probability is 0.75. In this way, a near-onset predicted as onset is penalized less than a false positive.
We also use data augmentation (DA) by randomly cropping each ROI from each sequence.
By combining DA and balancing, we obtain epochs with about 450,000 samples.
Finally, we manually use early-stopping to select the check-point to be evaluated (max.\ 15 epochs).

\section{Experimental testbed: Clarinetists for Science dataset}\label{sec:expsetup}
We acquired and annotated the new \emph{Clarinetists for Science} (C4S) dataset, released with this paper\footnote{For details, examples, and downloading see \url{http://mmc.tudelft.nl/users/alessio-bazzica\#C4S-dataset}}. C4S consists of 54 videos from 9 distinct clarinetists, each performing 3 different classical music pieces twice (4.5h in total). The videos have been recorder at 30 fps, about 36,000 events have been semi-automatically annotated and thoroughly checked.
We used a colored marker on the clarinet to facilitate visual annotation, and a green screen to allow for background augmentation in future work.
Besides ground-truth onsets, we include coordinates for face landmarks and 4 ROIs: mouth, left hand, right hand, and clarinet tip.

In our experiments, we use leave-one-subject-out cross validation to validate the generalization power across different musicians (9 splits in total).
From each split, we derive the 	training, validation and test sets from 7, 1, and 1 musicians respectively.
Hyper-parameters, like decaying learning rate and L2 regularization factors, are manually adjusted looking at f-scores and loss for train and validation sets.
We compute the f-scores using 50 ms as temporal tolerance to accept a predicted onset as true positive.
We compare to a ground-truth informed random baseline (correct number of onsets known) and to two state-of-the-art audio-only onset detectors (namely, SuperFlux \citep{Bock2013SuperFlux} and CNN-based \citep{DBLP:conf/icassp/SchluterB14}).

\section{Results and discussion}\label{sec:results}
During our preliminary experiments, most of the training trials were used to select optimization algorithm and suitable hyper-parameters.
Initially, gradients were vanishing, most of the neurons were inactive, and networks were only learning bias terms.
After finding hyper-parameters overcoming the aforementioned issues, we trained our model on 2 splits.

\begin{table}[h]
  \centering
  \begin{tabular}{| c || c | c || c |}
    \hline
    \textbf{method} & Split 1 & Split 2 & Average \\
    \hline
    informed random baseline & 27.4 & 19.6 & 23.5 \\
    \hline
    audio-only SuperFlux \citep{Bock2013SuperFlux} & 82.8 & 81.3 & 82.1 \\
    \hline
    audio-only CNN \citep{DBLP:conf/icassp/SchluterB14} & 94.3 & 92.1 & 93.2 \\
    \hline
    visual-based (proposed) & 26.3 & 25.0 & 25.7 \\
    \hline
  \end{tabular}
  \caption{F-scores with a temporal tolerance of 50 ms.}
  \label{tbl:results}
\end{table}

By inspecting the f-scores in \tabref{tbl:results}, we see that our method only performs slightly better than the baseline, and that the gap between audio-only and visual-based methods is large (60\% on average).
We investigated why and found that throughout the training, precision and recall often oscillate with a negative correlation.
This means that our model struggles with jointly optimizing those scores.
This issue could be alleviated by different near-onsets options or by formulating a regression problem instead of a binary classification one.

When we train on other splits, we observe initial f-scores not changing throughout the epochs.
We also observe different speeds at which the loss function converges.
The different behaviors across the splits may indicate that alternative initialization strategies should be considered and that the hyper-parameters are split-dependent.

\section{Conclusions}\label{sec:conclusions}
We have presented a novel cross-modal way to solve note onset detection visually.
In our preliminary experiments, we faced several challenges and learned that our model is highly sensitive to initialization, optimization algorithm and hyper-parameters.
Also, using a binary classification approach may prevent the joint optimization of precision and recall.
To allow further research, we release our novel fully-annotated C4S dataset.
Beyond visual onset detection, C4S data will also be useful for clarinet tracking, body pose estimation, and ancillary movement analysis.

\section*{Acknowledgments}
We thank the C4S clarinetists, Bochen Li, Sara Cazzanelli, Marijke Schaap, Ruud de Jong, dr.~Michael Riegler and the SURFsara Dutch National Cluster team for their support in enabling the experiments of this paper.

\bibliography{references}

\end{document}